\documentclass[letterpaper]{article}
\usepackage{flairs}
\usepackage{times}
\usepackage{helvet}
\usepackage{courier}
\usepackage{graphicx}
\usepackage{amsmath}
\begin{document}
%
\title{Leveraging Linguistic Coordination in Reranking N-Best Candidates For End-to-End Response Selection Using BERT}
\author{Mingzhi Yu, Diane Litman \\ University of Pittsburgh \\ Pittsburgh, PA\\ miy39@pitt.edu, dlitman@pitt.edu 
}
\maketitle
\begin{abstract}
\begin{quote}
Retrieval-based dialogue systems select the best response from many candidates. Although many state-of-the-art models have shown promising performance in dialogue response selection tasks, there is still quite a gap between R@1 and R@10 performance. To address this, we propose to leverage linguistic coordination (a phenomenon that individuals tend to develop similar linguistic behaviors in  conversation) to rerank the N-best candidates produced by BERT, a state-of-the-art pre-trained language model. Our results show an  improvement in R@1 compared to BERT baselines, demonstrating the utility of repairing machine-generated outputs by leveraging a linguistic theory. 
\end{quote}
\end{abstract}

\section{Introduction}
In recent years, end-to-end (E2E) dialogue systems have made remarkable progress. 
One important type of E2E system is the retrieval-based system \cite{chen2016enhanced,zhou2018multi}, which aims to select the best dialogue response from multiple candidates. The core of these systems is often formalized as a task to match dialogue context and response candidates.


The use of pre-trained language models has recently been attracting attention. With their rich contextualized input representation trained on a large amount of data, models such as ELMO \cite{peters2018deep}, XLNET \cite{yang2019xlnet}, and BERT \cite{devlin2018bert}
have achieved state-of-the-art performance on various NLP tasks, including the dialogue response selection task of DSTC8 track 2 \cite{kim2019eighth}.
Thus,  we also experiment with a pre-trained language model, namely BERT, as our baseline for  response selection. 

While prior work has shown promising response selection results, there is  still a gap between the recall at the best (R@1) and top 10 (R@10) candidates (see section Evaluation Metrics).
For many reported models on the leaderboard of DSTC 8 track 2, R@10 can achieve above 90\%, while R1 can be approximately 30\% lower on average. This gap between R@1 and R@10 indicates that the model failed to select the correct response, but the correct response is highly likely within the 10 best candidates. In the evaluation of our baseline model (see section Results), we also have a similar observation. 
This observation motivates us to rerank the N-best candidates (in our case, produced by BERT) with the goal of improving  model performance as reflected by R@1. 

Our approach is based on linguistic coordination, a phenomenon that individuals tend to linguistically mimic each other in the conversation. Previous research showed that language may converge in a wide range of linguistic dimensions such as lexical \cite{nenkova2008high,brennan1996lexical}, acoustic-prosodic \cite{rahimi2017entrainment,levitan2011measuring}, and linguistic styles \cite{gonzales2010language}. 
Therefore, we propose to leverage linguistic coordination, specifically lexical coordination to improve response selection.

\begin{figure}[]
\centering
\includegraphics[width=\columnwidth]{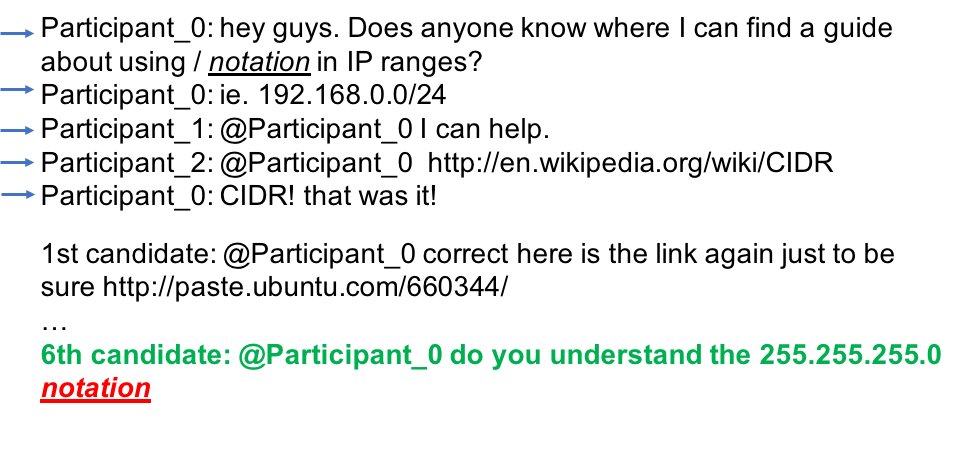}
\caption{A ranking example. Italic text shows possible coordination. The correct response is in bold text.}
\label{fig:example}
\end{figure}

Figure \ref{fig:example} shows example response ranks from a baseline model, where the correct answer is incorrectly ranked as the 6th rather than the top candidate.
In the correct response, the word ``notation" was mentioned in the context. This can signal higher lexical coordination. 
Examples such as this lead us to hypothesize that using linguistic coordination to rerank N-best candidates will improve R@1 performance.
To our knowledge, this is the first attempt to leverage linguistic coordination in the dialogue response selection task.
\section{Dataset}
We evaluate our approach on the Ubuntu IRC dataset \cite{acl19disentangle} provided by the eighth Dialog System Technology Challenge (DSTC8) track 2 subtask 1, which is the disentangled Ubuntu IRC dataset.
Each sample in the dataset consists of a multi-turn and multi-party dialogue with 100 response candidates that may be selected for the next turn. The ground-truth response might not be on the candidate list, leading to some no-answer cases. The DSTC 8 track 2 provides a train, development, and test set. A model is expected to rank the list of candidates by their possibility to be the utterance for the next turn. The evaluation metrics include Recall@k. Here the $k$ in the Recall@k means that the true positive response is among the first $k$ ranked candidates. 
Previous works reported Recall at 1, 2, 5 and 10 \cite{wu2020enhancing,Bertero2020model}. Intuitively, the higher value of a Recall@k indicates a stronger model capability to select the proper response. The smaller k is, the more robust the model is. 

We use all conversations from the train set for training. When a conversation has no answer, 
we use the last utterance in the context as the response and all prior utterances as the new dialogue context. There are two methods to perform evaluation considering there are some no-answer cases. The first method adds a no-answer candidate to the candidate list for each dialogue, and it treats the no-answer candidate as a response. The second method treats a dialogue as a no-answer case if the best candidate selected by the model has a confidence score lower than a certain threshold, e.g., 95\%. The method 
could potentially lead to an overestimated recall at higher values of $k$. For a naive example, by setting a high threshold for no-answer cases, e.g., 99.9\%, many dialogues in the test set can be predicted as no-answer cases. Thus, a model will put no-answer option at the top of its candidate ranking list as the best predicted candidate. Meanwhile, the remaining candidates may be kept the same order. This increases the odds of including the correct candidate in the ranking list predicted by the model. In the example of setting high threshed as 99.9\%, if this dialogue is a true no-answer case, then a model will score a hit for Recall@1, Recall@2 and Recall@10. On the opposite, if the dialogue is a false no-answer case, the model will miss the hit for Recall@1, but it is still highly likely to score a hit for Recall@2 and Recall@10. 

Therefore, for the simplicity of evaluation, we exclude no-answer cases from the development and test set, and we only evaluate on the dialogues with explicit answers. 
Note that the previous DSTC 7 \cite{yoshino2019dialog} provides similar datasets containing all having-answer conversations, but the conversations are two-party. 

We clean our dataset by removing dialogues containing empty candidates. In summary, we have 223,487 dialogues in the train set, 3,837 in the development set, and 4,384 in the test set.
Due to the nature of the Ubuntu dataset, there are many typos and technical terms such as urls and symbols. 
We limit the out-of-vocabulary words by preprocessing the datasets. Certain types of vocabulary are represented as abstract categories including paths, urls, symbols, file extensions, numbers, and addresses. We stemmed words.


\section{Methodology}
\subsection{Dialogue Modeling}
We follow previous works to formalize the response selection task as a supervised binary classification problem \cite{chen2016enhanced,zhou2018multi}. Each instance is a triple consisting of a label, dialogue context, and a response. The response in the positive case is the ground-truth response. The response in the negative case is randomly sampled from the remaining candidates. The ratio of positive to negative instance is treated as a hyper-parameter. Our ratio is 1:4. This results in approximately 1 million training cases.

\subsection{Baseline Model}
We use the BERT model as the baseline in a pre-training and fine-tuning manner. To adapt the dialogue to BERT model, we use the dialogue context and response as the two input segments suggested in generic BERT \cite{devlin2018bert}. Other configurations also follow the standard BERT for the sentence pairing task.
Similar to their sentence pair task model, we add a single fully-connected layer that takes the contextualized class representation $T_{cls}$ as the input and then a softmax layer to perform the binary classification task. We minimized the cross-entropy loss. The final output probability for the positive class will be used as the ranking score.

At the first pass, we will use the baseline BERT to generate the 10-best candidates. Then at the second pass, we will use our algorithm to rerank the 10-best candidates. Our evaluation will focus on the improvement introduced by leveraging linguistic coordination. While there are some existing works on the same dataset from DSTC 8 that are also based on BERT, we do not use them as the first-pass baseline models mainly because the models are not publicly available. 


\subsection{Reranking with Linguistic Coordination}
We focus on {\it lexical} coordination in this work. Inspired by \citeauthor{danescu2012echoes} \shortcite{danescu2012echoes}, we adapt their linguistic coordination measure to our specific problem and data. Compared with their original measure, we modify their formula to measure lexical overlap at the group-level. 
Coordination between a group and a member is found in other multiparty conversation datasets. Compared to a non-group member, a group member will have higher vocabulary overlap with the group \cite{rahimi2017entrainment}. Thus, we view the dialogue context and the response as a single turn exchange. Then we can adapt the measure to a group-level that measures coordination between a group and a member.   

Considering that we only have a turn exchange, we further modify the original formula to address the word coordination in a single exchange. Equation \ref{eq:Coor_m} shows the coordination, denoted as $Coor_m$, of a specific word $m$ in the context. Here $m$ functions as one lexical marker to evaluate lexical coordination. The $c$ and $r$ here denote the dialogue context and the current response. 
The minuend is 1 when $m$ is used in both the context and response. This indicates a word overlap occurs. Otherwise $Coor_m$ equals to 0. $P(\epsilon^{m}_{r})$ reflects how likely this word is used in the current data set. $K$ is a scalar that can be tuned accordingly. 

Equation \ref{eq:P_2} shows the calculation of $P(\epsilon^{m}_{r})$. $Count^{m}_{r}$ denotes the count of the marker $m$ in the current responses. $Count^{m}$ denotes the count of this marker m in the vocabulary count.  To ensure $P(\epsilon^{m}_{r})$ can reflect the word usage in the Ubuntu datasets, we create a larger vocabulary set using dialogues from the original task development set, as well as the top 10 candidates selected by our best baseline model. We ignore a union of stop words\footnote{From NLTK stop words dictionary}, a list of interjections, English numbers, and the most 200 common words based on the development set performance. 
In the case of negative $Coor_{m}$ value, we let $Coor_{m}$ be 0 so that the range of $Coor_{m}$ is between 0 and 1. In equation \ref{eq:score}, $Coor$ indicates the average coordination for all $m$ when $Coor_m$ is not 0.  The generic score $G$ is generated by the baseline model. The final score weights the generic score G and the average coordination score $Coor$. We tune the weights $w_g$ for G and $w_{coor}$ for $Coor$ on the development set. We bypass reranking if a candidate scores are very high in $G$ ($>$99\% based on our development set) because a high value in $G$ implies high confidence from the baseline model.




\begin{equation}
Coor_m =
\begin{cases}
    1 - K * P(\epsilon^{m}_{r}),     & \text{m in c and r}\\
    0,              & \text{otherwise}
\end{cases}
\label{eq:Coor_m}
\end{equation}



\begin{equation}
P(\epsilon^{m}_{r})  =  \frac{Count^{m}_{r}}{Count^{m}}
\label{eq:P_2}
\end{equation}



\begin{equation}
S  = w_g * G + w_{Coor} * Coor
\label{eq:score}
\end{equation}

\section{Experiments}

\subsection{Evaluation Metrics}
\label{sec:eval}
Our response selection evaluation metrics are Recall@1 (R@1) and Recall@10 (R@10). Here the $k$ in the Recall@k means that the true positive response is among the first $k$ ranked candidates. We also include Mean Reciprocal Rank (MRR) to measure the general ranking quality.  

\subsection{Baseline Implementation Details}
\label{sec:implementation}
We use the BERT TensorFlow implementation from Google. The pre-trained model is BERT-base uncased, which we fine-tune in 3 epochs. The maximum sequence length is set to 128. The batch size is 128. We train 2 models. The learning rate is $2e^-5$ and $1e^-5$ for Model 1 and 2, respectively. Dropout rate is 0.1. Models are trained on one GPU.

\subsection{Results and Discussion}
\label{sec:result}
We use Bert as the baseline and train two models, Model 1 (M1) and Model 2 (M2), to examine the robustness of our reranking method. Both use the same training configuration except for different learning rates, which results in M1 performing worse than M2 (Table \ref{tab:results}).

Table \ref{tab:results} shows the results. Overall, R1 is improved by reranking for M1 and M2. Note that R@10 maintains the same before and after reranking because we only rerank the top 10 candidates. For the development set, with the best weights between generic and coordination score, M1 shows the most improvement in R@1 by 6.31\%. MRR also increased by 4.22\%. M2 shows a small improvement of only 0.86\% in R@1. MRR increased by 0.63\%. Using the weights found in the development set, the test set results are similar. 
M1 also shows the most improvement in R@1 by 6.14\%. MRR increased by 4.21\%. M2 shows again shows a small improvement of 1.18\% in R@1. MRR increased by 0.68\%. Lexical coordination supports the baseline BERT models to further disambiguate optimal answers, especially for weaker models. However, as the results of baseline models improve, the utility of reranking becomes weaker.

\begin{table}[h]
\centering
\begin{tabular}{|l|cc|cc|}
\hline
 \multicolumn{5}{|c|}{Development Set}  \\ \hline
                          & \multicolumn{2}{c|}{Model 1}        & \multicolumn{2}{c|}{Model 2}     \\ \hline
                          & BERT                      & Rerank         & BERT    & Rerank         \\ 
R@1                       & 31.82                     & \textbf{38.13} & 41.93   & \textbf{42.79} \\ 
R@10                      & 64.22                     & 64.22          & 70.19   & 70.19          \\ 
MRR                      & 43.12                     & \textbf{47.34} & 51.65   & \textbf{52.28} \\ \hline
\multicolumn{5}{|c|}{Test Set}  \\ \hline
            & \multicolumn{2}{c|}{Model 1}               & \multicolumn{2}{c|}{Model 2} \\ \hline            
            & BERT                      & Rerank         & BERT    & Rerank         \\ 
R1          & 31.00                     & \textbf{37.14} & 39.74   & \textbf{40.92} \\ 
R10         & 64.10                     & 64.10          & 69.62   & 69.62          \\ 
MRR         & 42.26                     & \textbf{46.47} & 49.97   & \textbf{50.65} \\ \hline
\end{tabular}
\caption{The evaluation results for the development and test set. The recall is shown in percentage.}
\label{tab:results}
\end{table}

Table \ref{tab:test_error_position} shows the correct response distribution at the top three positions before and after reranking for the test set. For M1 we observe an increase in the 1st position but a decrease in both the 2nd and 3rd positions. One possible explanation is that correct response are initially ranked very closely to the 1st position, and the subsequent reranking slightly adjusts the order based on coordination. Similarly to M1, M2 also shows an increase in the 1st position, but it also shows an increase in the 3rd position. The result implies that some cases are incorrectly reranked. 


\begin{table}[h]
\centering
\begin{tabular}{|c|ll|ll|}
\hline
 & \multicolumn{2}{c|}{Model 1}               & \multicolumn{2}{c|}{Model 2} \\ \hline
Ranking & BERT & Rerank  & BERT & Rerank          \\ 
1        &31.00 & \textbf{37.14} & 39.74 & \textbf{40.92}\\ 
2       &9.58  & 7.50 & 9.08 & 8.00 \\ 
3       & 5.68 & 4.47 & 5.06 & \textbf{5.13}\\ \hline
\end{tabular}
\caption{The correct answer position in the test set}
\label{tab:test_error_position}
\end{table}

\subsection{Case Study}
We analyze the reranking outputs. Here error cases are those baseline cases that failed to rank the correct response as first place. For the test set, Table \ref{tab:analysis} shows the number of error cases that have linguistic coordination in the correct response (and thus potentially could have been corrected by our reranking method), as well as the number of actual corrections and errors after reranking. The caps are 11.13\% and 5.82\% for M1 and M2 respectively. The higher percentage for M1 indicates that for the weaker model, there is more space for reranking to improve.
\begin{table}[h]
\centering
\begin{tabular}{|l|l|l|l|}
\hline
     &                                 & Model 1                      & Model 2  \\ \hline
     & Cap                             & 488                     & 255 \\ 
Test Set & Correction                      & 360                     & 102 \\ 
     & New Error                       & 91                      & 50  \\ \hline
\end{tabular}
\caption{Test set case analysis. Cap: \# of error cases having coordination in the best candidate. Correction: \# of corrections after reranking. New Error: \# of new errors after reranking}
\label{tab:analysis}
\end{table}

\begin{figure}[h]
\centering
\includegraphics[width=\columnwidth]{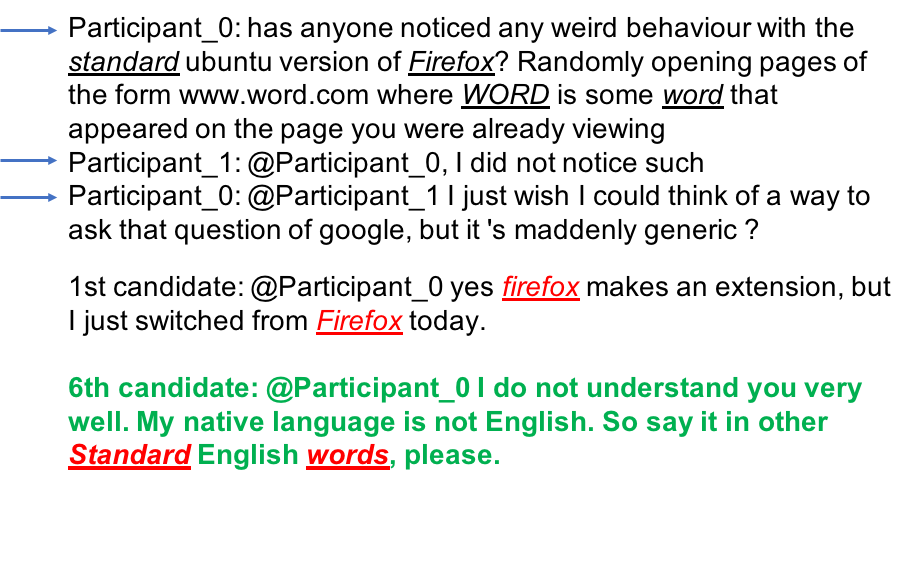}
\caption{A correction example from the test set. Italic text shows potential lexical coordination. Bold text shows the correct response.}
\label{fig:correction_case}
\end{figure}

Figure \ref{fig:correction_case} is a typical example of correction from selecting a low-coordinate response to a high-coordinate response. The correct response is in the 6th position. Both the 1st candidate and 6th candidates show some coordination at `firefox', `standard', and `word'. The word `standard' and `words' are used less frequently in the dataset but they are both used in the 6th candidate. Thus, the score of the 6th candidate is in higher coordination than the first candidate. This implies that the reranker can capture the coordination in word usage beyond counting word repetition. This is also an interesting example in that the correct response is not directly related to the context but contains context words.

\section{Conclusion and Future Work}
We proposed a simple yet effective approach to rerank the N-best candidates generated by a pre-trained language model. The results show promising improvement in selecting the correct response. By leveraging linguistic coordination, this work provides a way to rerank the response candidates in an unsupervised manner. 
Our approach does not increase the computational complexity, and can be applied when there is limited access to computational resources. In the future, we intend to use other state-of-the-art models in the second-pass reranking and experiment on more datasets.

\bibliography{main}
\bibliographystyle{flairs}

\end{document}